\documentclass{article}

\usepackage{arxiv}

\usepackage[utf8]{inputenc} 
\usepackage[T1]{fontenc}    
\usepackage{hyperref}       
\usepackage{url}            
\usepackage{booktabs}       
\usepackage{amsfonts}       
\usepackage{nicefrac}       
\usepackage{microtype}      
\usepackage{xcolor}         
\usepackage{comment}
\usepackage{graphicx}
\usepackage{bbm}
\usepackage{tikz}
\usepackage{etoolbox}
\usepackage{pifont}
\usepackage{xspace}
\usepackage{subcaption}
\usepackage{adjustbox}
\usepackage{tabularx}
\usepackage{algorithm}
\usepackage{algorithmic}
\usepackage{amsmath}
\usepackage{graphicx}
\usepackage{xspace}

\graphicspath{ {./figs/} }

\usetikzlibrary{arrows.meta,
                calc,
                positioning,
                quotes,
                automata}

\newcommand{\graphs}{\mathcal{G} \xspace}
\newcommand{\graphsn}{\tilde{\mathcal{G}}\xspace}
\newcommand{\edges}{\mathcal{E} \xspace}

\newcommand{\spotlight}{\textrm{SL}}
\newcommand{\mset}{\mathcal{M}\xspace}
\newcommand{\gset}{\mathcal{G}\xspace}
\newcommand{\gsim}{\textrm{sim}_G\xspace}
\newcommand{\algoname}{\texttt{MotiFiesta}\xspace}
\newcommand{\spminer}{\texttt{SP-Miner}\xspace}
\newcommand{\micrograph}{\texttt{Micro-Graph}\xspace}
\newcommand{\REAFUM}{\texttt{REAFUM}\xspace}
\newcommand{\RAM}{\texttt{RAM}\xspace}
\newcommand{\VEAM}{\texttt{VEAM}\xspace}
\newcommand{\mfinder}{\texttt{Mfinder}\xspace}

\def\Z{{\mathbf Z}}
\def\z{{\mathbf z}}
\def\Y{{\mathbf Y}}

\title{Approximate Network Motif Mining via Graph Learning}

\author{
  Carlos Oliver\\
  Department of Biosystems Science and Engineering \\
  ETH Z\"urich \\
  \texttt{carlos.oliver@bsse.ethz.ch} \\
  \And
  Dexiong Chen\\
  Department of Biosystems Science and Engineering \\
  ETH Z\"urich\\
  \url{dexiong.chen@bsse.ethz.ch}
  \AND
  Vincent Mallet\\
  Structural Bioinformatics Unit, Department of Structural Biology and Chemistry,\\
  Institut Pasteur, Universite de Paris-Cité, CNRS UMR3528 \\
  Center for Computational Biology, Mines ParisTech,  Paris-Sciences-et-Lettres Research University \\
  \url{vincent.mallet96@gmail.com}
  \And
  Pericles Philippopoulos \\
  Ozeki Inc.\\
  \url{pericles@ozeki.io}
  \And
  Karsten Borgwardt\\
  Department of Biosystems Science and Engineering \\
  ETH Z\"urich\\
  \url{karsten.borgwardt@bsse.ethz.ch}
}

\begin{document}
\maketitle


\begin{abstract}    
Frequent and structurally related subgraphs, also known as network motifs, are valuable features of many graph datasets. 
However, the high computational complexity of identifying motif sets in arbitrary datasets (motif mining) has limited their use in many real-world datasets. 
By automatically leveraging statistical properties of datasets, machine learning approaches have shown promise in several tasks  with combinatorial complexity and are therefore a promising candidate for network motif mining. 
In this work we seek to facilitate the development of machine learning approaches aimed at motif mining.
We propose a formulation of the motif mining problem as a node labeling task.
In addition, we build benchmark datasets and evaluation metrics which test the ability of models to capture different aspects of motif discovery such as motif number, size, topology, and scarcity. 
Next, we propose \algoname, a first attempt at solving this problem in a fully differentiable manner with promising results on challenging baselines. 
Finally, we demonstrate through \algoname that this learning setting can be applied simultaneously to general-purpose data mining and interpretable feature extraction for graph classification tasks.
\newline

\end{abstract}

\section{Introduction}


The hypothesis motivating network motif mining is that substructures that play an important role in the network are more frequent than expected.
A common approach for uncovering meaningful substructures in graph datasets is thus to search for sets of over-represented subgraphs, a.k.a. network motif mining.
This was first illustrated when Milo et al. ~\cite{milo2002network} proposed \mfinder performed an exhaustive search for occurrences of 3-node subgraphs occurring with a higher frequency than expected in network datasets from various domains.
The resulting motifs have since been studied extensively and yielded insights in the functioning of systems such as cellular signalling and ecological networks ~\cite{doncic2013feedforward,rip2010experimental}, confirming the utility of motif mining. 
More recently, several learning algorithms have demonstrated that once identified, network motifs can be leveraged in many ways. 
As building blocks, network motifs have been shown to assist for realistic graph generation in molecules ~\cite{jin2020hierarchical}.
Besta et al. ~\cite{besta2021motif} propose the task of predicting whether a set of nodes will form a specified motif as a challenging training regime for representation learning.
Motifs have also been proposed as explanations of black box classification models ~\cite{perotti2022graphshap}.
Despite exhaustive network motif mining being NP-hard ~\cite{yu2020motif}, the utility of motif libraries has made the task of motif discovery a key challenge in data mining for the past 30 years.
The aim of this work is to formally expose motif mining to machine learning models in an effort to enhance both the discovery of motifs and the representation power of learned models.


Any motif mining algorithm has to perform two computationally intensive steps: subgraph search and graph matching.
The process of discovering a new occurrence of the motif involves a search over the set of subgraphs of a given dataset which yields a search space that grows exponentially with the number of nodes in the dataset and in the motif ~\cite{hartmanis1982computers}. 
Next, for a given candidate motif and subgraph, a graph matching procedure known to belong to NP ~\cite{conte2004thirty} is needed to determine whether the candidate can be included in the set of instances of the motif.
Despite these barriers, many motif mining tools have been proposed, all of which rely on simplifications of the task or {\it a priori} assumptions about the desired motifs .
These simplifications include bounding motif size, number of motifs, topology constraints, and simplified subgraph matching criteria. 
Moreover, real world network datasets often represent dynamic or noisy processes, making their motifs approximate. 
Examples of approximate motifs have been observed in protein and RNA molecules which exhibit a large degree of physical flexibility and genetic variability ~\cite{oliver2022vernal,reid2010variable}, noise in image data ~\cite{acosta2012frequent}, and medical time series features ~\cite{bock2018association}.
Algorithms which limit their matching procedure to exact isomorphism will overlook a large set of possible motif candidates ~\cite{schreiber2010motifs}.

For this reason, we emphasize that our proposed methodology is built to support the discovery of \emph{approximate} motifs. 
That is, we wish to allow non-isomorphic subgraphs to participate in the same motif class given sufficient structural similarity.
Some tools have addressed this challenge, again with strict limitations such as \VEAM ~\cite{acosta2012frequent}and \REAFUM \cite{li2015reafum} which allow small motifs to vary in node and edge labelling, and \RAM ~\cite{ram} which tolerates a fixed number of edge deletions within a given motif.
To our knowledge, none of the approximate motif mining tools applicable in a general data setting or have available implementations.


The recent success of graph representation learning, particularly in an unsupervised context, presents an opportunity for circumventing some of these bottlenecks ~\cite{karalias2020erdos}.
Namely, by allowing graph embedding models to leverage the statistical properties of a dataset, we can cast the search and matching problems to efficient operations such as real-valued vector distances. 
This idea was first explores in the specific case of RNA structural motifs ~\cite{oliver2022vernal} but has not been tested on general graphs.
Continuing this intuition, some recent machine learning methods have been proposed for problems related to motif mining.
\spminer ~\cite{spminer} detects frequent subgraphs by building an \emph{order embeddings} which efficiently encodes subgraph-supergraph relationships in a graph dataset. 
The order embedding lets a decoder efficiently identify subgraphs with many occurrences but motifs occurrences are restricted to those isomorphic with respect to the Weisfeiler-Lehman test (not approximate), and not endowed with a notion of statistical over-representation.
Finally, \micrograph ~\cite{zhang2020motif} learns a lookup table which represents a fixed number of subgraphs useful for graph-subgraph contrastive learning.
While not guaranteeing that subgraphs fulfill formal motif properties, \micrograph highlights the view of motifs as feature sets which casts motif mining not just as an end for network analysis but also a means towards stronger graph neural networks. 
Taking this idea further, this work aims to directly tackle the problem of approximate motif mining such that the full potential of learning methods is leveraged.

\subsection{Contributions}

In this work, we (1) formalize the notion of approximate motif mining as a machine learning task and provide appropriate evaluation metrics as well as benchmarking datasets, (2) propose \algoname, a fully differentiable model as a first solution to learnable approximate motif mining
which discovers new motifs in seconds for large datasets, and (3) we show that motif mining could also serve as an effective unsupervised pre-training routine and interpretable feature selector in real-world datasets. 

\section{Approximate Motif Mining Task}

Our first contribution is a definition of the approximate motif mining, and a general framework for evaluating motif mining models. 

\textbf{Network Motif Definition}. Given a collection of graphs $\gset = \{ G_1 , G_2, .., G_M \}$, with each $G_i = (V_i, E_i, \bf X_i)$ being a tuple of node and edge sets with a $d$-dimensional node feature matrix, a
 network motif is a subgraph $g \subset \gset$ fulfilling the following two properties: 
\begin{enumerate}
     \item \textbf{Connectivity:} Every $g$ is connected (i.e. $\exists (p \sim q) \forall (p, q) \in N \times N$). 
     \item \textbf{Concentration: } $\frac{|\mset(g | \gset)|}{|\mset(g | \tilde{\gset})|} > c$,  for concentration some threshold $c \in \mathbb{R}^+$, occurrence set \\ $\mset(g | \gset) = \{g' : s_G(g, g') > b \quad \forall g' \subset \gset \}$, graph similarity function $\gsim : \gset \times \gset \rightarrow [0, 1]$, and similarity threshold $b \in [0, 1]$.
\end{enumerate}

The first criterion ensures that motifs represent coherent structural subunits and is often applied in real-world datasets.
For example, when working with protein graphs, structural subunits ( e.g. domains) are known occupy a convex volume which is reflected in graphs as a locally connected subgraph.
The concentration condition ensures that $g$ is significantly over-represented by computing the cardinality of the set of graphs related to $g$, $|\mathcal{M}(g)|$.
To determine statistical significance, a set of null graphs $\tilde{\gset}$ is built by a random process which preserves generic statistics of $\gset$ but removes any bias towards specific substructures.
This step allows us to filter out spurious motifs, although omitting $\tilde{\gset}$ defines the task of frequent subgraph mining ~\cite{spminer}.
Typically, $\tilde{\gset}$ is constructed by randomizing $\gset$ for example by swapping pairs of edges such that the degree distribution and number of nodes are preserved ~\cite{milo2002network}.
For the classical motif mining problem, $\gsim$ can be chosen to yield 1 iff $g \simeq g'$ and 0 otherwise. 
An approximate mining method allows for some degree of discrepancy $b < 1$ and thus has a better chance at improved recall when $\gset$ is noisy or models a dynamic system. 
Often $\gsim$ is implemented as a graph kernel ~\cite{oliver2022vernal,vishwanathan2010graph} or an edit distance ~\cite{li2015reafum,riesen2009approximate}.

The full motif mining task is thus to identify all $g \subset \gset$  that satisfy the the above conditions.

\subsection{Motif mining as node assignment}

Assuming an algorithm exists that identifies all motif-like graphs, evaluating whether the proposed subgraphs fulfill the concentration criterion would still be intractable as it would involve expensive search and matching routines.
Instead, we propose as in ~\cite{spminer} to build a synthetic dataset where motifs are artificially  injected at high concentrations at known positions.
In this context, and in real-world datasets where ground-truth motifs are known, evaluating the performance of a motif miner can be done through efficient set comparisons. 

\begin{figure}
    \centering
    \includegraphics[width=\textwidth]{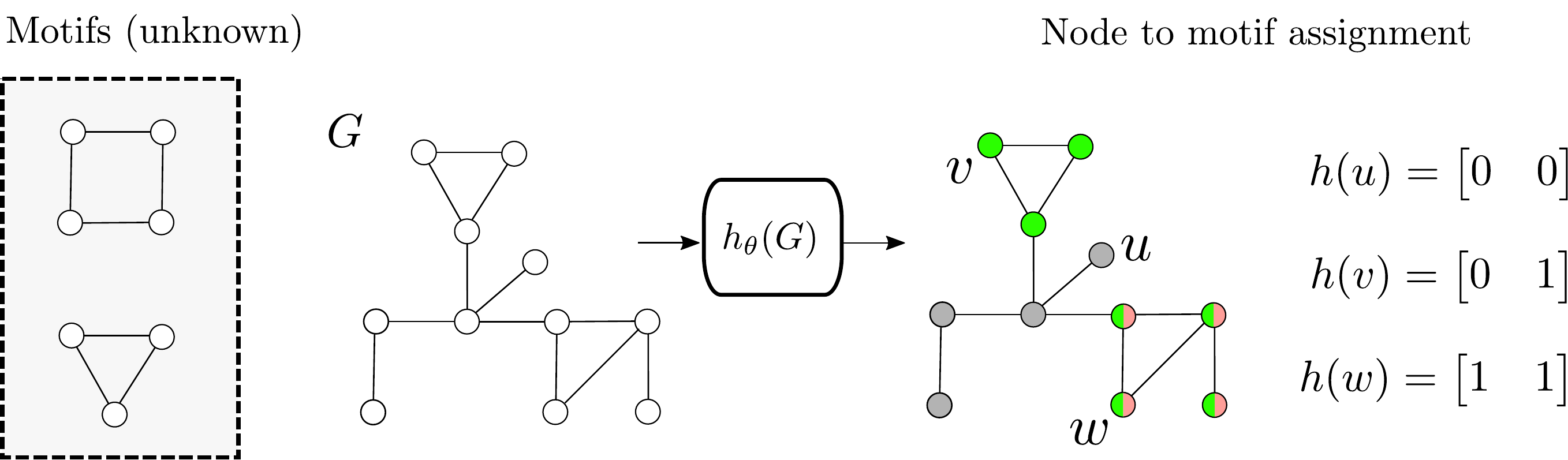}
    \caption{Illustration of the approximate motif mining task. Model $h_\theta$ computes a $K$-dimensional (2 in this case) assignment for each node in the input graph, without access to ground truth motifs. Ground truth motif nodes are colored with the same color and used at test time only.}
    \label{fig:approx_task}
\end{figure}

More precisely, we say a motif mining model is a function $h_{\theta}: G \rightarrow [0, 1]^{|V| \times K}$ parametrized by $\theta$ which maps a graph to a $K$-dimensional node labeling.
We denote the (soft) labeling with $\hat{\Y}$ which specifies the probability that node $i$ belongs to motif $j$.
In this setting, the number of motifs $K$ is known \emph{a priori} but is not given to the model at training time.
Note that this framing allows for soft (probabilistic) assignment of nodes to motifs, non-isomorphic motif occurrences, as well as one node belonging to multiple motifs (non-disjoint motifs).
\textbf{Figure ~\ref{fig:approx_task}} illustrates this procssess.

Graph datasets represent systems whose behaviour is often governed by complex interaction patterns. 

When considering a single dimension (motif),  we see that $\hat{\Y}_{*,j}$ partitions $\gset$ into those nodes inside motif $j$, $\hat{\Y}_{*,j} \rightarrow 1$ and those outside the motif $\hat{\Y}_{*, j} \rightarrow 0$.
At this point, we wish to measure the agreement between two sets for all motifs, for which the Jaccard coefficient is widely accepted.
Here, we use the generalization of the Jaccard coefficient to real-valued sets. 
To evaluate the model applied to a single graph, $\hat{\Y} = h_\theta(G)$ we have,

\begin{equation}
    \textrm{J}(\hat{\Y}, \Y) = \frac{1}{K}\sum_{j=1}^{K} \frac{\sum_{i=1}^{|V|} \min (\Y_{i,j}, \hat{\Y}_{i,j})}{\sum_{i=1}^{|V|} \max(\Y_{ij}, \hat{\Y}_{ij})}, \text{where} \quad \hat{\Y}, \Y \in [0, 1]^{|V| \times K}.
\end{equation}
The Jaccard coefficient ranges from 0 to 1 and simultaneously captures precision and recall.
Models which over-assign nodes to a given motif are penalized by the denominator which considers the union of the prediction and the true motif, while a model that misses motif nodes will have a low numerator value. 

Because there is no inherent ordering to motif sets, the labeling assigned by the model and the one chosen for ground-truth motifs will be arbitrary.
The final performance measure for a motif labeling is therefore given by the maximum Jaccard coefficient over all permutations $S_K$ of the columns of $\Y$:
\begin{equation}
    \textrm{M-Jaccard}(\Y, \hat{\Y}) = \max_{\pi \in S_K} \textrm{J}(\hat{\Y}, \pi (\Y)).
\end{equation}

When motif datasets are constructed following the motif criteria defined above, or when motif locations are known in real-world datasets, the M-Jaccard indirectly but efficiently measures the ability of a model at capturing motif-likeness.
The permutation term limits $K$ to small values, which leads to the hypothesis that models able to show strong performance on small $K$ will yield valid motifs in general.
Testing this hypothesis would require expensive graph matching procedure between two sets of subgraphs (predicted vs known motifs).

\section{\algoname: Approximate Network Motif Mining Model} 

Here, we propose a model that tackles the approximate motif mining problem. 
The focus of our algorithm, \algoname is to train a model that fulfills the connectivity condition, and directly learns substructure concentrations.
We do so through a combination of edge-based pooling, substructure representation learning, and frequency estimation layers.
The main idea behind \algoname is to leverages the well-known modular and hierarchical nature of motifs whereby larger motifs are composed of smaller ones ~\cite{schreiber2010motifs,reinharz2018mining}.
This principle allows the model to speed up the subgraph search component of motif mining.
More specifically, we apply the intuition that an edge between two motifs has the potential to give rise to a larger motif.

\subsection{Learnable edge contraction}

To bring this idea into a differentiable framework we adapt the $\textrm{EdgePool}$ layer proposed in ~\cite{diehl2019edge} as our basic operation. 
For an input graph, the $\textrm{EdgePool}$ operation collapses pairs of adjacent nodes into a single node according to a learned probability over their node embeddings. 
When a pair of nodes is collapsed, the newly created node is an embedding for the new node is computed as a function of the two original node embeddings.
The resulting graph contains  a new set of nodes and node features, each of which contain pooled information from a pairs of nodes in the input graph.
If this process is repeated, the newly computed nodes contain information from larger and larger subgraphs in the original graph.
Moreover, because all pooling events are carried out over edges, we guarantee that the subgraph being represented is always connected which inherently satisfies the first criteria of motifs.
See Appendix for more details on the EdgePool implementation.

In \algoname we view a node in the coarsening graph as representing a subgraph in the original graph, and the contraction operation as growing a motif. 
We define the spotlight of node $\spotlight(u)$, as the subgraph in G whose nodes have been coarsened into $u$.
Before the first iteration of contraction $ \spotlight(u) = \{u\} \quad \forall u \in V$. 
Then for each coarsening event, a pair of connected nodes is collapsed into a new node $w$ and so we have $\spotlight(w) = \spotlight(u) \cup \spotlight(v)$.
Taking this $\textrm{EdgePool}$ operation and spotlights as basic elements, we train the model to fulfill the structural similarity and concentration criteria of motifs through a two loss objective.
We outline the execution of the learning process in \textbf{Algorithm ~\ref{algo:train}}.

\begin{algorithm}[H]
\caption{\algoname learnable edge contraction}\label{algo:train}
\begin{algorithmic}[1]
    \STATE {\bfseries Input:} a batch of graphs $\graphs$ , $\graphsn$, graph similarity function $\gsim$, density estimator $\hat{f}$
    \STATE {\bfseries Output:} embedding model $\phi$ and scoring model $\sigma$ 
    \STATE  $SL(u) \leftarrow \{u\}$ for each $u \in \graphs$ tracks receptive field of coarsened nodes
    \STATE $\graphs^{(0)} \leftarrow \graphs$
    \FOR{$t=1,\dots,T$}
        \STATE $\graphs^{(t)} \leftarrow \emptyset$ \\
        \FOR{$(u, v) \in \edges^{(t)}$}
            \STATE Compute joint embedding $\z_{uv} \leftarrow \phi(u, v)$\\
            \STATE Compute joint negative embedding $\tilde{\z}_{uv} \leftarrow \phi(u', v')$\\
            \STATE Compute edge score $s(u, v) \leftarrow \sigma(z_{uv})$
            \STATE Add node $w$ to $\graphs^{(t)}$ w.p. $s(u,v)$ \\
            \STATE Update spotlights $\spotlight(w) \leftarrow \spotlight(u) \cup \spotlight(v)$
        \ENDFOR
    \STATE Backprop $\mathcal{L}_{rep}(\phi) \leftarrow || \langle z_u, z_v \rangle - \gsim(\spotlight(u), \spotlight(v)) ||_2^2$ for all nodes. 
    \STATE Backprop $\mathcal{L}_{conc}(s) \leftarrow -s(u, v) \cdot \exp{[-\beta \Delta_{f}(\z; \Z, \tilde{\Z})]}$ for all edges. 
    \ENDFOR
\end{algorithmic}
\end{algorithm}


\subsection{Substructure representation loss}

The first loss term of our model ensures that similarities in node embeddings reflect structural similarities of the subgraphs (spotlights) they represent.
This step is necessary for ensuring that subsequent density estimates in the embedding space point to regions of high structural similarity.
Given any graph similarity function,  $\gsim$, and a graph embedding function $\phi: G \rightarrow \mathbb{R}^{|V| \times d}$, we define the \textit{representation} loss for a pair of nodes $(u, v)$ as:
\begin{equation}
    \mathcal{L}_{rep}(u, v, \phi) = || \langle \phi(u), \phi(v)\rangle  - \gsim(\spotlight(u), \spotlight(v)) ||_2^2.
\end{equation}
At each layer, the embedding function is applied to all adjacent nodes $(u, v)$ to obtain an embedding that represents the union of subgraphs contained in $\spotlight(u)$ and $\spotlight(v)$. 
The choice of similarity function will dictate the implicit rule determining motif belonging and thus application specific considerations can be encoded at this step ~\cite{oliver2022vernal}. 
That is, by choosing $\gsim$, we are implicitly determining the subgraphs admitted into $\mset(g)$,
Additionally, we are not restricted to functions with explicit feature maps or proper kernels and can use the inductive nature of $\phi$ to avoid evaluating similarity across all possible pairs of nodes.
For these experiments we use the Wasserstein Weisfeiler Leman graph kernel ~\cite{togninalli2019wasserstein} which jointly models structural similarity and node feature agreement on general undirected graphs.


\subsection{Concentration loss}

The embedding model $\phi$ assigns similar embeddings to spotlights with similar structures. 
Next, we need to identify which of these edge-induced subgraphs fits the concentration criterion of motifs.
We therefore learn a function $\sigma: \mathbb{R}^{d} \rightarrow [0, 1]$ which assigns a probability $s_{uv}$ to each subgraph such that highly concentrated subgraphs have a a large probability and vice versa.
We implement $\sigma$ with a simple MLP and train it using an efficient density estimator $\hat{f} : \mathbb{R}^{d} \rightarrow \mathbb{R}^+$ so that the second loss term for a batch of edge-induced spotlights $Z$ and a vector of spotlight scores, $\sigma$ is given by
\begin{equation*}
\mathcal{L}(\sigma, \Z, \tilde{\Z} ; \beta, \lambda) = -\sum_{i=1}^{|\Z|} \sigma_i \exp[-\beta \Delta_f(\z_i; \Z, \tilde{\Z})] + \lambda ||{\mathbf \sigma}||^2_2,
\end{equation*}
where $\Delta_f(\z_i; \Z, \tilde{\Z}) = \hat{f}_{\Z}(\z_i) - \hat{f}_{\tilde{Z}}(\z_i)$, $\lambda$ is the regularization strength, and $\beta$ controls the growth of $\sigma$ for more concentrated motifs.
The delta function measures the difference in concentration of the given subgraph under the input graph distribution against its density in a randomized graphs with embedding matrix $\tilde{\Z}$.
To generate $\tilde{\Z}$ we apply $\phi$ to $\gset$ using the randomization technique described in ~\cite{schreiber2010motifs} which iteratively swaps pairs of edges with each other such that local connectivity patterns are disrupted but the global graph statistics such as size and degree distribution are maintained.

Embeddings with a high density under the input graphs and low denesity under the randomized distribution will result in a large negative contribution to the loss.
In this case a $\sigma_i = 1$ will best minimize the loss.
Sparsely concentrated subgraphs, or those with a high concentration in both distributions will have their $\sigma$ pushed to zero by the regularization term. 
As we expect motifs to occur less frequently than other embeddings the loss function is likely unbalanced.
For this reason we apply an exponential to the delta function so that non-motif embeddings don't overwhelm the loss.

There are several choices for the density estimate $\hat{f}$, given some sample of embeddings $\Z$, but here we choose the $k$-NN density estimator ~\cite{mack1979multivariate} for its efficiency.
We let $\hat{f}_{\Z, k}(\z) \sim \frac{k}{N} \times \frac{1}{V^{d} R_{k}(\z;\Z)}$ where $R_{k}(\z;\Z)$ is the distance between $z$ and its $k$-th nearest neighbor in $\Z$ which defines the radius of a $d$-dimensional sphere of volume $V^{d} = \frac{\pi^{d/2}}{\Gamma(d/2 + 1)}$ and $\Gamma(z)$ is the Gamma function.
The intuition for this estimator is that points falling into dense regions will have many points nearby and result in smaller $k$-nearest neighbor spheres.

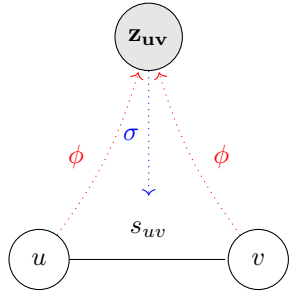
\begin{figure}[h]
    \centering
     \begin{tikzpicture}[shorten >=1pt,auto,node distance=2.1cm, minimum size=.8cm]
\node[draw,circle] (u) {$u$};
\node[draw,circle,right=of u] (v) {$v$};
\node[draw,circle,fill=black!10, above=of $(u.north)!0.5!(v.north)$] (z) {$\bf z_{uv}$};
\draw (u) edge node[above](s){$s_{uv}$} (v);
\path[->] (u)  edge[red, dotted,bend right=10] node [left] {$\phi$} (z);
\path[->] (v)  edge[red, dotted, bend left=10] node[right] {$\phi$} (z);
\path[->] (z)  edge [blue, dotted] node[left,xshift=5pt] {$\sigma$} (s);
\end{tikzpicture}
    \caption{The representation model $\phi$ computes an embedding $z_{uv}$ for the subgraph induced by $(u, v)$ which $\sigma$ uses to assign a merging probability.}
    \label{fig:variables}
\end{figure}

The combined flow of both models is shown in \textbf{Figure ~\ref{fig:variables}}.

\subsection{Decoding}

The output required to evaluate $\textrm{M-Jaccard}$ coefficient is an assignment of each node to a fixed number of discrete categories (motifs).
However, the model does not assign embeddings to any categories, instead $\sigma_i$ as a scalar only models the motif-likeness of a given subgraph.
To obtain an explicit assignment we apply a locality sensitive hash ~\cite{lsh} to all the embeddings in the dataset.
Each embedding is assigned to a bucket by a locality sensitive hashing function such that similar embeddings are assigned to the same bucket with high probability, leveraging the similarity properties enforced by our representation loss.
The bucket ID can then be taken as an integer code of the nodes in the original graph, yielding a full partitioning of the graphs and an assignment of nodes to motifs.
Nodes are either assigned a motif ID or no motif assignment depending on a threshold applied to the subgraph score $\sigma_i$.
At this point we apply the $\textrm{M-Jaccard}$ coefficient permutation test when we have a ground truth motif labeling.
This process is summarized in \textbf{Algorithm ~\ref{algo:decode}}.

\begin{algorithm}
\caption{\algoname motif decoding}\label{algo:decode}
\begin{algorithmic}[1]
    \STATE {\bfseries Input:} a batch of graphs $\graphs$, motif mining $h_\theta$, contraction level $T$
    \STATE {\bfseries Output:} node to motif assignment matrix $\hat{Y}$ 
    \STATE ${\mathbf Z}, {\mathbf S }\leftarrow m(\graphs, T)$ collect all node embeddings and $\sigma$ scores
    \STATE $H \leftarrow \textrm{HASH}({\mathbf Z})$ map each embedding to an integer code.
    \STATE Let $\mathcal{H}$ contain the set of unique hash codes in $H$.
    \STATE $\hat{\Y} \leftarrow \textrm{OneHot}(H, \mathcal{H})$
    \STATE Let $\textrm{Rank}_{\mathcal{H};S}$ return the rank of a hash code $h$ by descending mean score in ${\mathbf S}$.
    \FOR{$h \in \mathcal{H}$}
        \IF{$\textrm{Rank}_{\mathcal{H}, {\mathbf S}}(h) > r$}
            \STATE $\hat{\Y}[h] \leftarrow \bf 0$
        \ENDIF    
    \ENDFOR
    \RETURN{$\hat{\Y}$}
\end{algorithmic}
\end{algorithm}


\subsection{Complexity analysis}

The bulk of runtime is spent on the training step , particularly the subgraph similarity learning stage, which is performed once per dataset.
Since we chose the Wasserstein Weisfeiler Lehman graph kernel, evaluating the representation loss for $m$ subgraphs requires $\mathcal{O}(m^2)$ calls to the WWL kernel with runtime $\mathcal{O}(n^3\log(n))$ for $n$-node subgraphs.
The density estimation is built on a nearest neighbor estimator which relies on quick neighbourhood searches which can be implemented on a K-d tree in $\mathcal{O}(\log(m))$.
In the induction phase, we execute a forward pass through the scoring and pooling which is $\mathcal{O}(m)$.
To discretize the subgraphs we use an $\mathcal{O}(n)$ locality sensitive hashing procedure.
We note that polynomial and sub-polynomial runtimes for classical enumeration methods are very rare.
Typical training times on were around 2 hours for the synthetic datasets on a single GPU, and the decoding step takes $\sim$ 10 seconds. 
\section{Results}

We test the proposed model and evaluation framework in three settings.
The first experiment tests the ability of \algoname to retrieve motifs in synthetic datasets using when the ground truth motifs are known using the M-Jaccard coefficient.
Next, we explore the potential for the motif mining task as an unsupervised pre-training step in graph classification setting.
Finally, to explore the relevance and interpretability of the motif mining procedure we perform an ablation study on our mined motifs to search for important motifs during classification.

\subsection{Mining for motifs in synthetic datasets}

To evaluate motif mining models with ground truth and in a controlled setting, we generate several synthetic datasets that capture different motif-related variables.
Each dataset consists of 1000 graphs generated randomly using the Erd\"os-Reyni random graph generator.
Next, we generate one or more subgraphs that will act as the motifs.
To create concentrated subgraphs we insert the motifs graphs into the each of the original graph and randomly connect nodes in the motif to the rest of the graph.
In this manner we are able to know which nodes belong to motifs and can control the concentration, size, topology, and structural variability of the motifs.
More details on dataset construction can be found in Appendix.

\begin{figure*}
    \centering
    \begin{subfigure}[b]{0.45\textwidth}
        \centering
        \includegraphics[width=\textwidth]{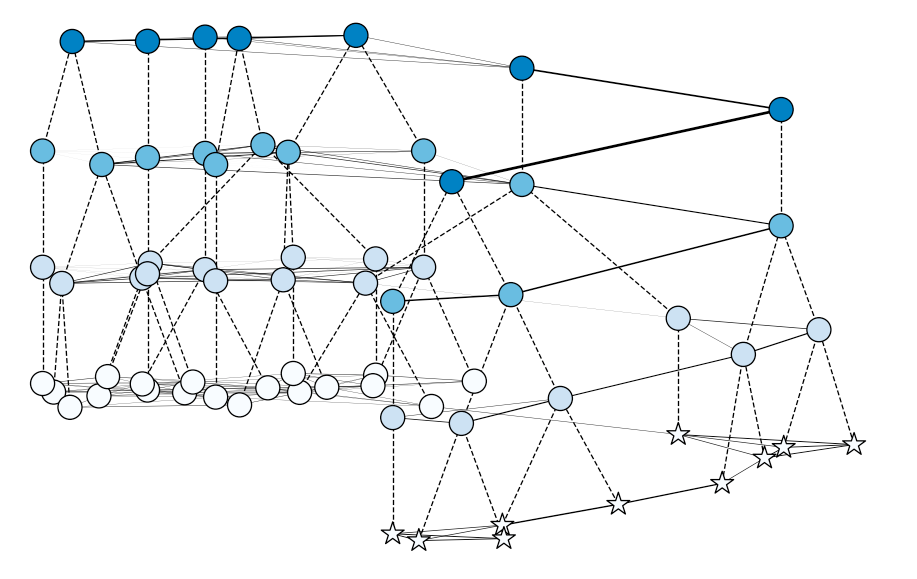}
        \caption{Illustration of edge pooling execution.}
        \label{fig:merging}
    \end{subfigure}%
    ~
    \begin{subfigure}[b]{0.45\textwidth}
        \centering
        \includegraphics[width=\textwidth]{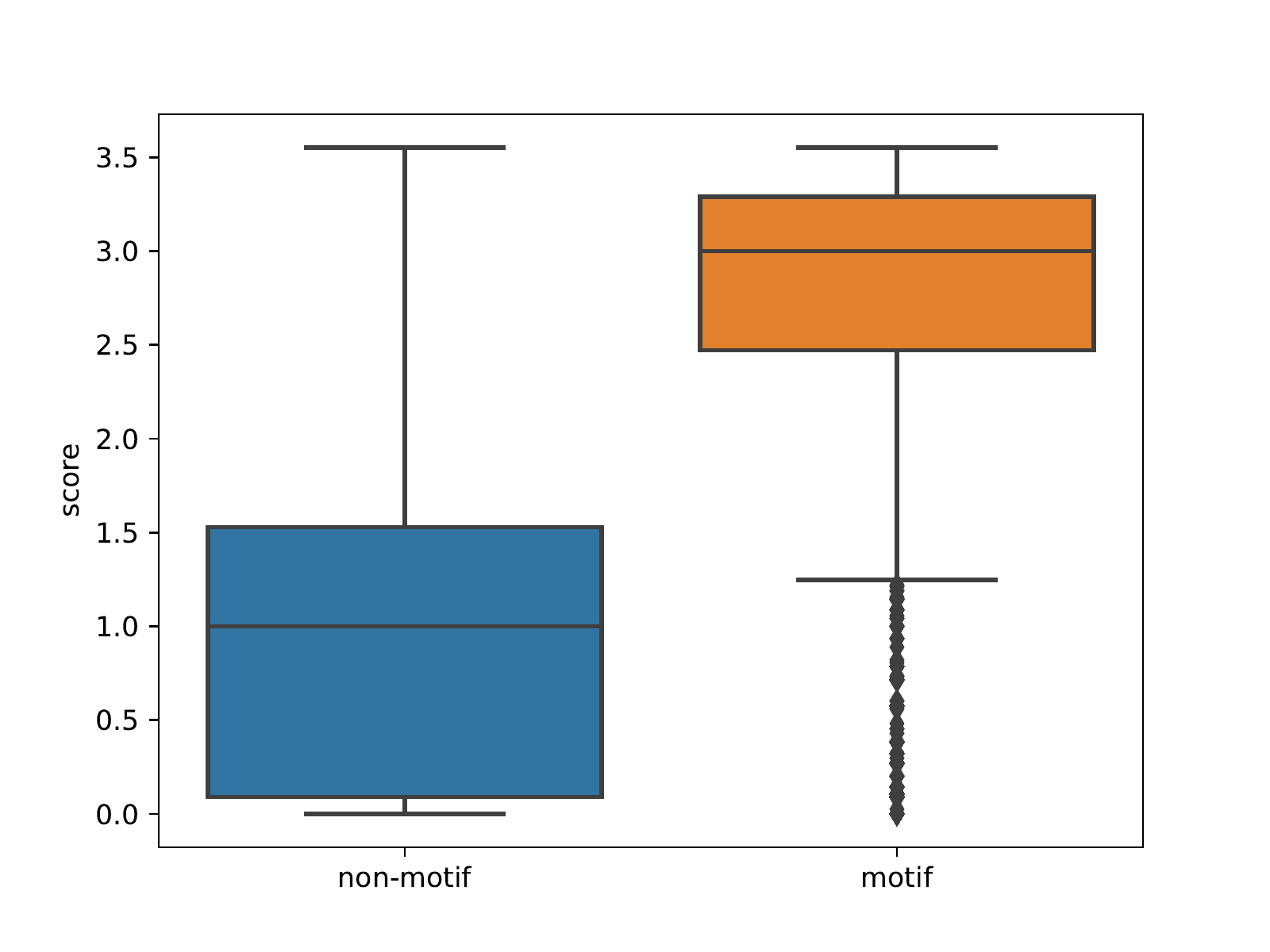}
        \caption{Pooling score distribution.}
        \label{fig:sigmas}
    \end{subfigure}
    \caption{(a) input graph is at the bottom with white nodes, and motif nodes drawn with a star shape. Pooled nodes are connected to the collapsed node by a dotted line. Illustration shows stacked 3 pooling layers. (b) distribution of edge scores assigned to nodes within a known motif versus those outside of the motif subgraph.}
\end{figure*}

After training \algoname on a synthetic dataset, we compute the $\textrm{M-Jaccard}$ coefficient both on the original data distribution, as well as on datasets with increasing degrees of distortion applied to the motif subgraphs.
Before applying the decoding step, however we can already assess the quality of the subgraph scoring layer.
That is, knowing which nodes belong to motifs, we should expect that measuring $\sigma$ should be significantly larger in nodes belonging to motifs than for non-motif nodes.
In \textbf{Figure ~\ref{fig:sigmas}}  see a clear separation between the two distributions, indicating that the model is able to assign proper scores to motif subgraphs.
As a visual aid, we show the pooling process on an example graph in \textbf{Figure ~\ref{fig:merging}}.

\begin{figure*}[htbp]
    \centering
    \begin{subfigure}[b]{0.33\textwidth}
        \centering
        \includegraphics[width=\textwidth]{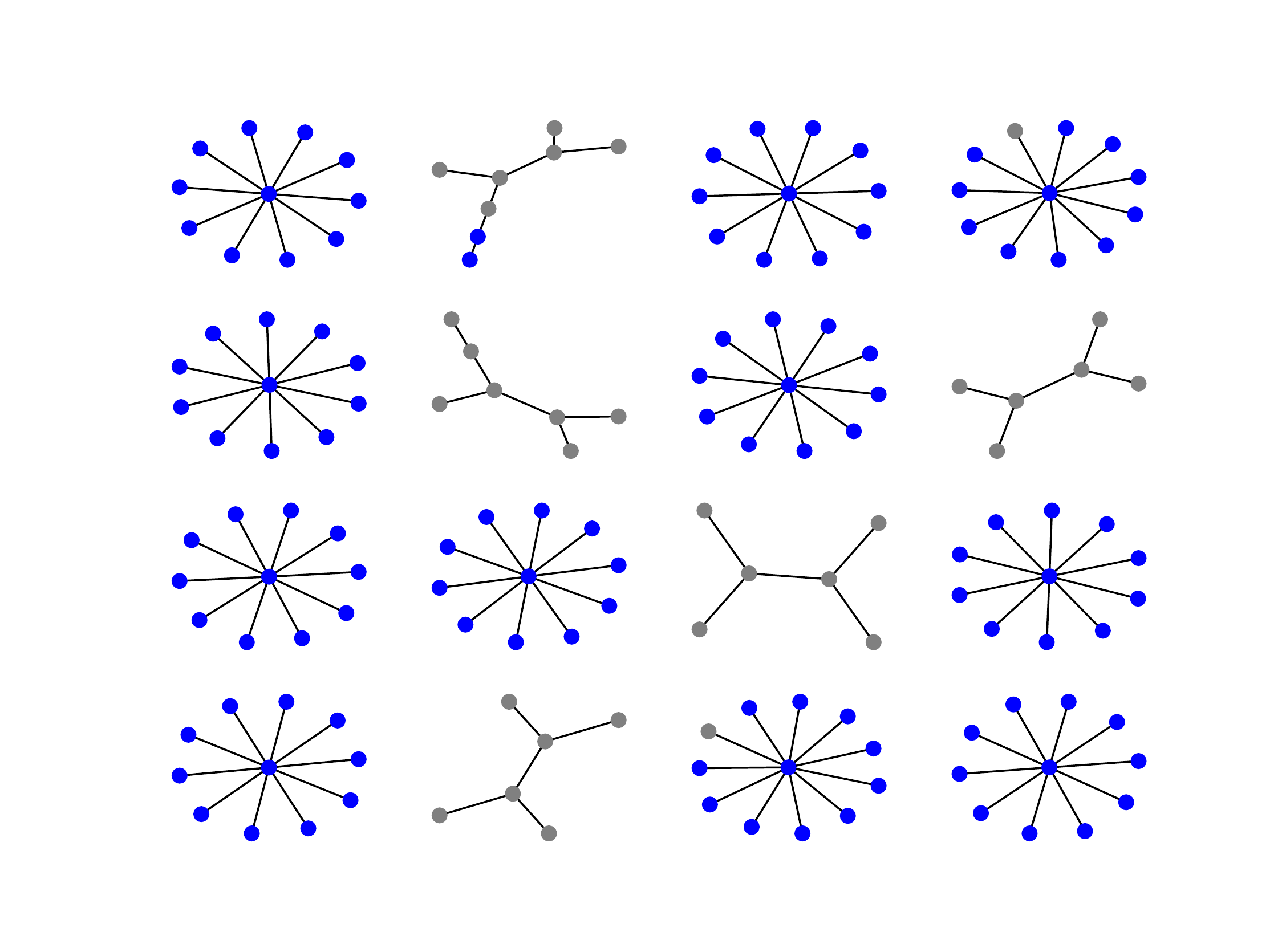}
        \caption{Star}
    \end{subfigure}%
    ~
    \begin{subfigure}[b]{0.33\textwidth}
        \centering
        \includegraphics[width=\textwidth]{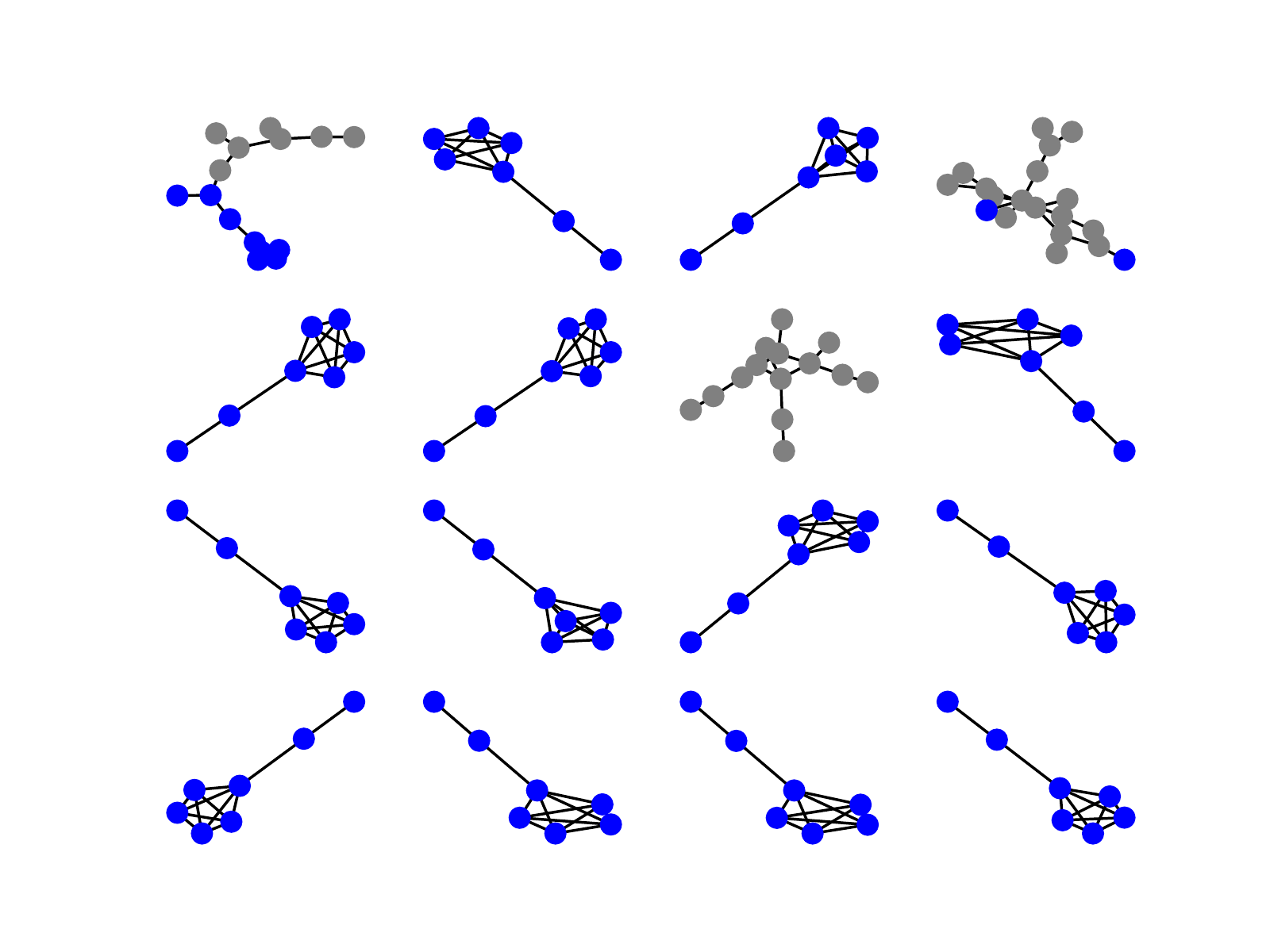}
        \caption{Barbell}
    \end{subfigure}
    \begin{subfigure}[b]{0.31\textwidth}
        \centering
        \includegraphics[width=\textwidth]{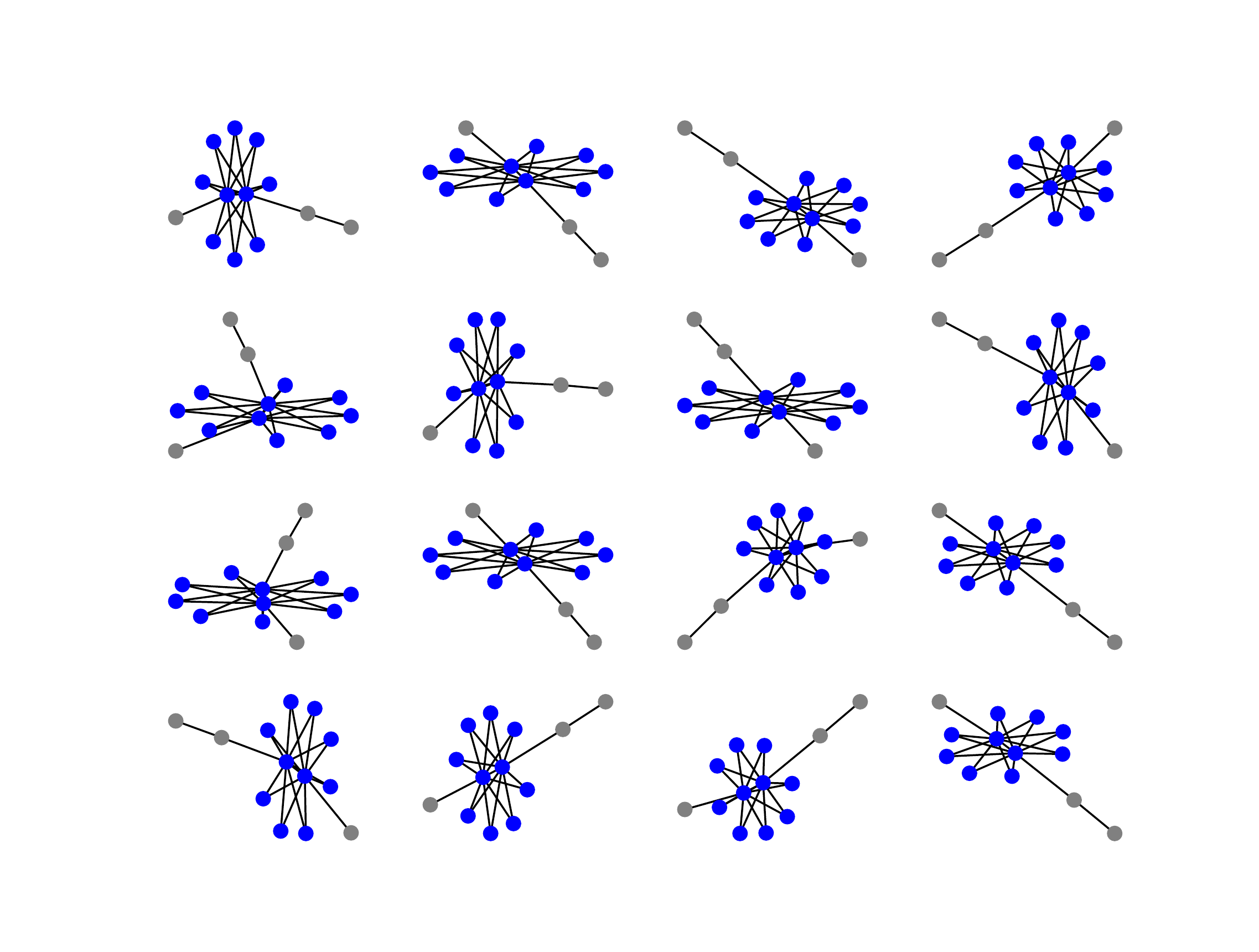}
        \caption{Clique}
    \end{subfigure}    
    \caption{Random samples from most populated buckets with ground truth motif nodes in blue.}
    \label{fig:examples}
\end{figure*}

Having checked that \algoname merging probabilities behave as expected, we apply the decoding phase to create final motif assignments for each node in the input graph and compute the $\textrm{M-Jaccard}$ coefficient.
We test three main motif mining conditions: motif topology (barbell, clique, star, and random), number of motifs (1, 3, 10) in the dataset, and varying true concentration of the motif (100\%, 50\%, 30\%).
As a control condition, for each dataset we train an equivalent dummy model with the only difference being that $\sigma_i = 0.5$ for all nodes.
The hyperparameter set remains the same for all experiments (see Appendix). 
The decoding step uses a small grid search over the flexibility parameter in the hashing step (hash size) as well as in the pooling layer to use (subgraph size), we output the max average M-Jaccard over 5 repetitions for each condition.
\textbf{Table ~\ref{table:jaccs}} shows that \algoname is able to consistently outperform the baseline in many conditions.
Notably, it appears that when the topology of the motif is well-defined (as in the case of the barbell and clique) we see the strongest results, indicating that the power of the graph embedding model is an important factor.
The lowest performances were seen in cases where motif occurrences are rare (sparse) and in the case of multiple motifs (3 and 5 motifs).
It is likely that further hyperparameter selection will be necessary to improve performance in these cases, we leave this to future work.
As a demonstration, we visualize randomly selected subgraphs from the top scoring bucket for the different motif topologies in \textbf{Figure ~\ref{fig:examples}}.

\begin{table}
\centering
\begin{adjustbox}{width=1\textwidth}
\begin{tabular}{lllll}
\toprule
{} & \multicolumn{4}{l}{} \\
 &                                        $\epsilon=0$ &                                        $\epsilon=0.01$ &                                        $\epsilon=0.02$ &                                        $\epsilon=0.05$ \\
\textbf{} &                                             &                                             &                                             &                                             \\
\midrule
\textbf{barbell} &  0.68 $\pm$ 0.04 {\small (0.40 $\pm$ 0.01)} &  0.66 $\pm$ 0.05 {\small (0.37 $\pm$ 0.02)} &  0.69 $\pm$ 0.01 {\small (0.32 $\pm$ 0.05)} &  0.67 $\pm$ 0.02 {\small (0.34 $\pm$ 0.06)} \\
\textbf{clique } &  0.58 $\pm$ 0.14 {\small (0.35 $\pm$ 0.00)} &  0.62 $\pm$ 0.08 {\small (0.34 $\pm$ 0.00)} &  0.57 $\pm$ 0.15 {\small (0.34 $\pm$ 0.00)} &  0.41 $\pm$ 0.08 {\small (0.34 $\pm$ 0.00)} \\
\textbf{random } &  0.51 $\pm$ 0.03 {\small (0.34 $\pm$ 0.00)} &  0.49 $\pm$ 0.02 {\small (0.35 $\pm$ 0.00)} &  0.46 $\pm$ 0.02 {\small (0.35 $\pm$ 0.00)} &  0.44 $\pm$ 0.04 {\small (0.35 $\pm$ 0.00)} \\
\textbf{star   } &  0.43 $\pm$ 0.09 {\small (0.43 $\pm$ 0.03)} &  0.42 $\pm$ 0.02 {\small (0.40 $\pm$ 0.04)} &  0.40 $\pm$ 0.02 {\small (0.38 $\pm$ 0.04)} &  0.40 $\pm$ 0.03 {\small (0.38 $\pm$ 0.03)} \\
\midrule
\textbf{5 nodes         } &  0.37 $\pm$ 0.02 {\small (0.34 $\pm$ 0.00)} &  0.36 $\pm$ 0.00 {\small (0.35 $\pm$ 0.01)} &  0.36 $\pm$ 0.01 {\small (0.36 $\pm$ 0.00)} &  0.36 $\pm$ 0.00 {\small (0.35 $\pm$ 0.02)} \\
\textbf{10 nodes        } &  0.48 $\pm$ 0.02 {\small (0.35 $\pm$ 0.00)} &  0.46 $\pm$ 0.03 {\small (0.35 $\pm$ 0.00)} &  0.43 $\pm$ 0.01 {\small (0.38 $\pm$ 0.04)} &  0.47 $\pm$ 0.03 {\small (0.35 $\pm$ 0.01)} \\
\textbf{20 nodes        } &  0.48 $\pm$ 0.12 {\small (0.34 $\pm$ 0.01)} &  0.43 $\pm$ 0.08 {\small (0.35 $\pm$ 0.01)} &  0.44 $\pm$ 0.03 {\small (0.39 $\pm$ 0.05)} &  0.33 $\pm$ 0.00 {\small (0.34 $\pm$ 0.00)} \\
\midrule
\textbf{3 motifs        } &  0.19 $\pm$ 0.01 {\small (0.18 $\pm$ 0.01)} &  0.29 $\pm$ 0.05 {\small (0.18 $\pm$ 0.01)} &  0.24 $\pm$ 0.06 {\small (0.17 $\pm$ 0.00)} &  0.29 $\pm$ 0.06 {\small (0.19 $\pm$ 0.01)} \\
\textbf{5 motifs        } &  0.11 $\pm$ 0.00 {\small (0.11 $\pm$ 0.00)} &  0.11 $\pm$ 0.00 {\small (0.11 $\pm$ 0.00)} &  0.11 $\pm$ 0.00 {\small (0.11 $\pm$ 0.00)} &  0.11 $\pm$ 0.00 {\small (0.11 $\pm$ 0.00)} \\
\midrule
\textbf{sparse-0.30  } &  0.36 $\pm$ 0.00 {\small (0.33 $\pm$ 0.02)} &  0.36 $\pm$ 0.01 {\small (0.35 $\pm$ 0.00)} &  0.35 $\pm$ 0.01 {\small (0.34 $\pm$ 0.01)} &  0.35 $\pm$ 0.00 {\small (0.34 $\pm$ 0.00)} \\
\textbf{sparse-0.50  } &  0.40 $\pm$ 0.03 {\small (0.34 $\pm$ 0.00)} &  0.43 $\pm$ 0.02 {\small (0.34 $\pm$ 0.00)} &  0.40 $\pm$ 0.04 {\small (0.34 $\pm$ 0.00)} &  0.43 $\pm$ 0.05 {\small (0.34 $\pm$ 0.01)} \\
\textbf{sparse-1.00  } &  0.40 $\pm$ 0.02 {\small (0.34 $\pm$ 0.00)} &  0.42 $\pm$ 0.03 {\small (0.32 $\pm$ 0.02)} &  0.41 $\pm$ 0.02 {\small (0.34 $\pm$ 0.00)} &  0.42 $\pm$ 0.04 {\small (0.34 $\pm$ 0.00)} \\
\bottomrule
\end{tabular}
\end{adjustbox}
\caption{$\textrm{M-Jaccard}$ score under various synthetic motif mining conditions. Each row contains the recovery score when testing on motif datasets with each edge of the motifs distorted with probability $\epsilon$. The values in small type and parentheses are the scores obtained by a dummy model that assigns all edges a merging probability of $0.5$.}
\label{table:jaccs}
\end{table}

\subsection{Motif mining as pre-training for graph classification}

Apart from applications in data mining, we explore the potential for the motif task as a pre-training phase for supervised learning.
We therefore test the view that motifs are statistically robust and structurally complex feature sets.
Taking three real world datasets in the TUDataset ~\cite{tu} from various applications domains we train \algoname using only the representation and concentration loss again with rewired graphs as a negative distribution.
Once the motif mining model is trained we apply a global pooling to each merging layer and concatenate the output of each layer to produce a graph embedding so that for $T$ pooling layers of $d$ dimensions each, we obtain a $T \times d$ feature vector for each graph.
The motif model is then frozen and we train a random forest classifier for each dataset taking as input the pooled embeddings.
As baselines we also train three graph neural network architectures in an end to end fashion (EdgePool, GCN ~\cite{kipf2016semi}, GIN~\cite{xu2018powerful}).
Since our baseline models are trained in an end to end fashion and our graph embeddings are instead fixed and sent to a linear classifier we do not expect \algoname to outperform the baselines. 
Rather, comparable performance against a model trained in an end to end fasion would point towards the motif mining task as a promising pre-training step.
Indeed, in \textbf{Table ~\ref{table:supervised}} we see that across three different datasets, using motif-based embeddings has similar performance to state of the art graph neural network architectures.

\begin{table}[htbp]
\centering
\begin{tabular}{lccc}
\toprule
             & PROTEINS &   COX2 & IMDB-BINARY \\
\midrule
\textbf{MotiFiesta} & 73.1$\pm$2.0  & 80.7$\pm$2.4 &  72.2$\pm$3.3  \\
\textbf{EdgePool} \cite{diehl2019edge} & 73.6$\pm$4.1 & 80.5$\pm$4.0 & 71.8$\pm$3.6 \\
\textbf{GCN} \cite{kipf2016semi}& 73.5$\pm$5.6 & 80.9$\pm$4.0 & 72.8$\pm$3.1 \\
\textbf{GIN} \cite{xu2018powerful}& 72.6$\pm$3.9 & 79.6$\pm$5.1 & 73.4$\pm$3.2 \\
\bottomrule
\end{tabular}
\caption{Classification performance of \algoname on graph classification benchmarks.}
\label{table:supervised}
\end{table}


\subsection{Motifs interpretability}

In real-world datasets, we lack ground-truth motifs, making definite motif mining evaluation difficult.
Despite this limitation, we propose to study the importance of the mined motifs through ablation studies on classification tasks. 
Our intuition is that if the motifs identified in real-world datasets hold important information about the functioning of the network, excluding their representations should have an impact on classification performance.
More specifically, when building the global embedding for a graph, we pool together spotlight embeddings at each layer.
In the previous experiment, we simply take all spotlight embeddings to obtain a graph embedding.
In this experiment, we only take spotlights with the top $k$ $\sigma_i$ scores.
In this manner, only the subgraphs that the pre-trained model deems most motif-like are allowed to contribute to the global graph embedding. 
After this process, we repeat the same classification tasks as in \textbf{Table ~\ref{table:supervised}} with varying $k$.
As a control for the differing amount of information entering the global embedding with varying $k$, we also traina a model where $k$ spotlights are chosen at random. 
As expected, we see a general upward trend in performance as $k$ grows while the top $k$ filtered models outperform the random model for low values of $k$. 
This indicates that in a fully unsupervised manner, the model shows promise for identifying useful subgraphs.
As a final step in interpretability, 
one could then simply inspect motifs with large $\sigma$ values for potentially functional subgraphs.
And owing to the reconstruction term in the training, all similar subgraphs are also given for free which would be a valuable asset for domain experts.
We leave detailed inspection of these subgraphs for future work.

\begin{figure*}[h!]
    \centering
    \begin{subfigure}[b]{0.45\textwidth}
        \centering
        \includegraphics[width=\textwidth]{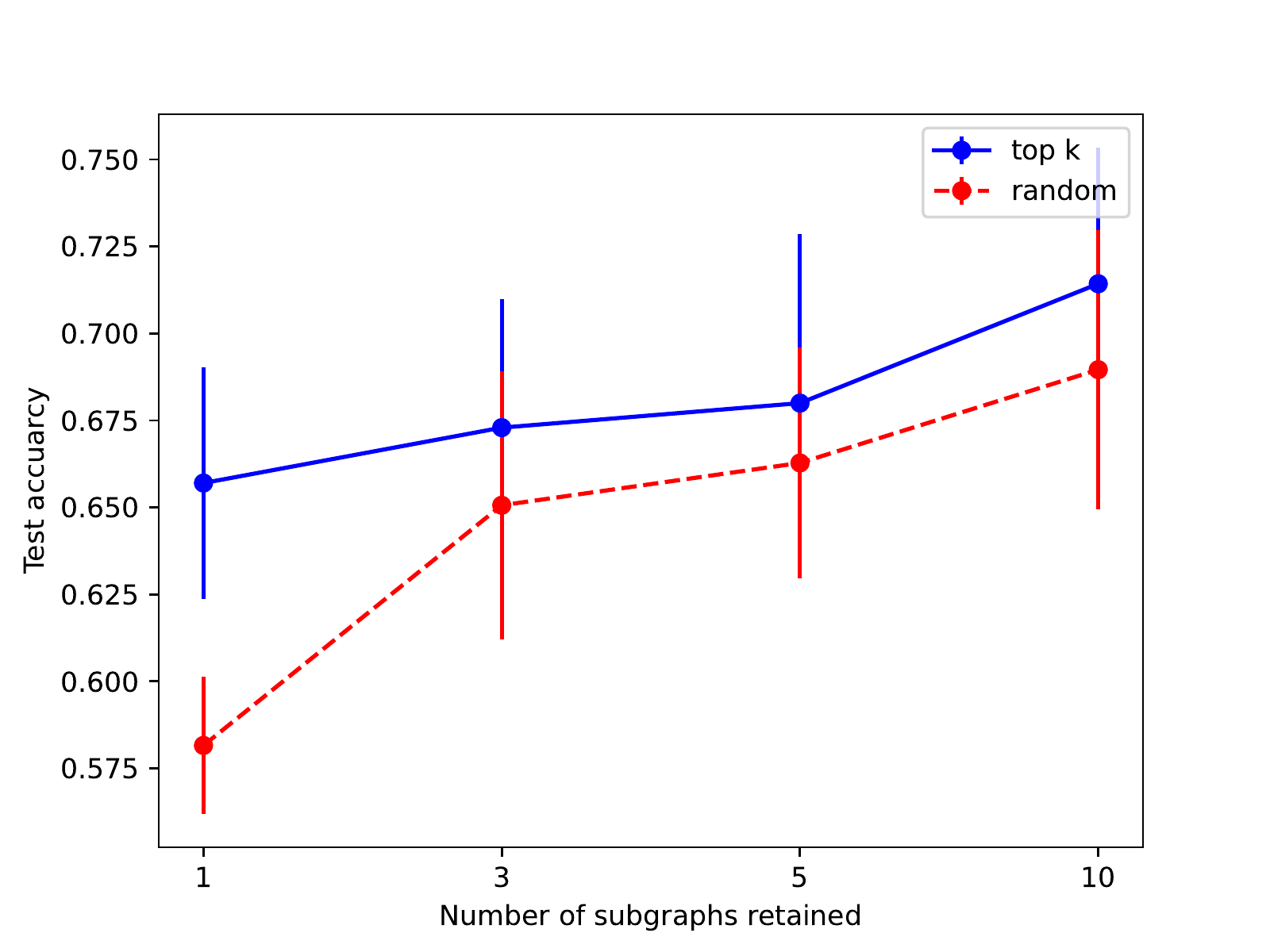}
        \caption{PROTEINS}
    \end{subfigure}%
    ~
    \begin{subfigure}[b]{0.45\textwidth}
        \centering
        \includegraphics[width=\textwidth]{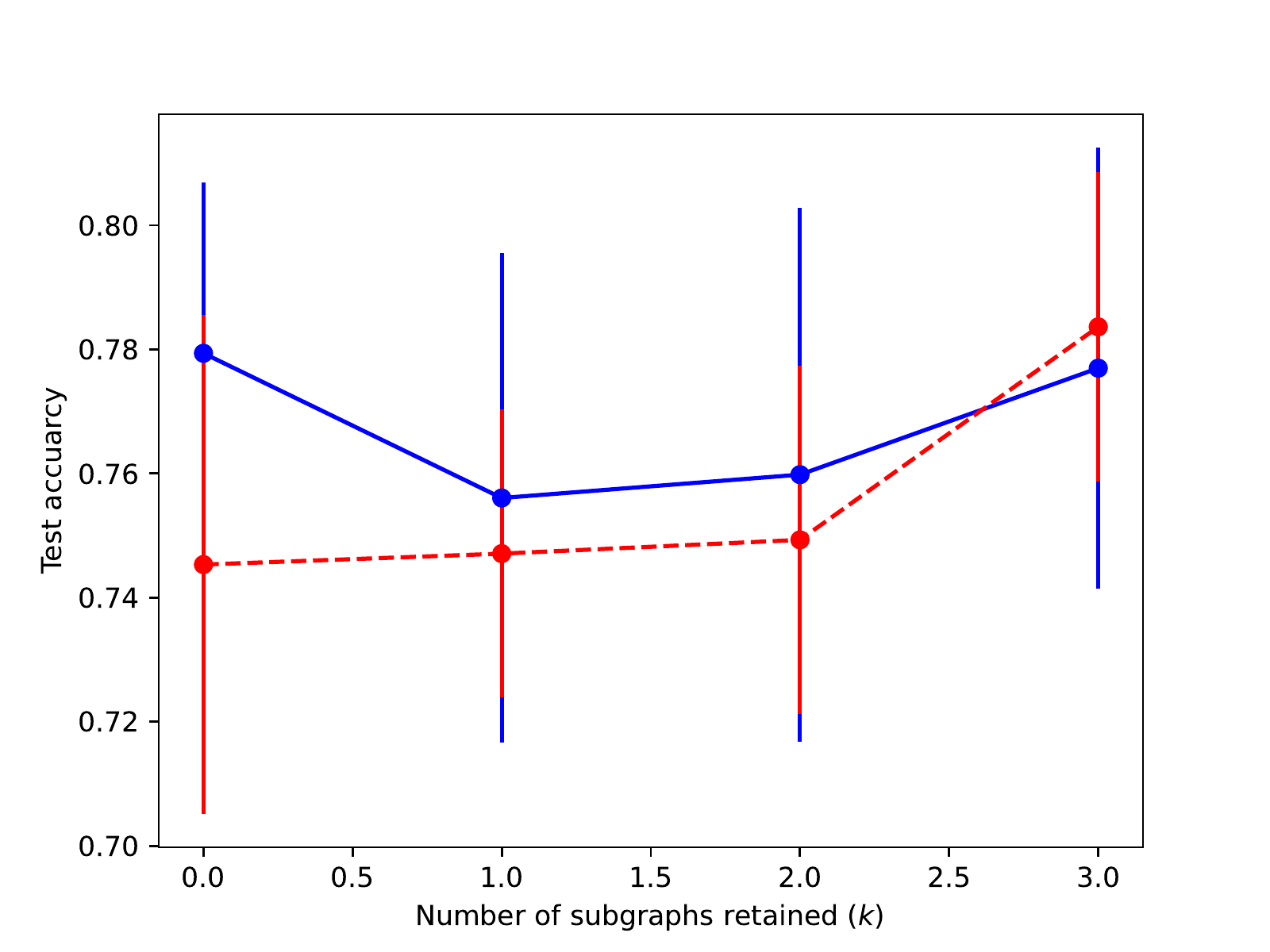}
        \caption{COX2}
    \end{subfigure}
    \caption{Subgraph ablation study}
    \label{fig:ablate}
\end{figure*}
\section{Conclusion}

We introduced a framing of the approximate network motif problem in a manner suitable for machine learning models, and propose a first model architecture to address this challenge. 
The model we propose for this task is able to efficiently recover motifs over baselines in several synthetic data conditions with decoding times on the order of seconds.
When placed in a classification setting we show that motif mining has potential as a challenging pre-training step and that the obtained motifs show potential to be naturally used as interpretable and robust feature extractors. 
Although the inductive nature of our model provides a significant performance boost at inference time, the subgraph similarity loss computation is a quadratic of a polynomial time kernel  which could limit the sensitivity of the model.
Additionally, the decoding phase requires tuning of several parameters which has to be carefully chosen for each application.
Further exploration of of efficient subgraph representation functions and graph partition models will therefore become relevant topics for motif mining.

\section{Availability}
\label{sec:avail}
Code and data for all experiments is available at \url{https://github.com/BorgwardtLab/MotiFiesta}.


\bibliographystyle{plain}
\bibliography{biblio}


\newpage
\appendix
 \section{Appendix}

The appendix contains additional details regarding dataset preparation, benchmarking, and experiment setup. 

\subsection{Dataset preparation}

To test motif mining models on datasets with ground truth motifs we generate synthetic graphs and repeatedly insert motifs to create over-represented subgraphs at known positions.
The dataset construction process admits several choices including number and type of motifs to insert, abundance and degree of distortion across motif instances.
For a given motif graph $m$ and randomly generated graph $G$.
We randomly sample a node from $G$ to delete and replace with $m$.
Each $G$ is generated using the Erdos-Reiny random graph generator with a connection probability $p=0.1$, and twice as many nodes as $m$.
Next, we randomly sample pairs of nodes with one coming from $m$ and the other from $G$ as the set of edges that connect $m$ to the rest of $g$ using the same edge probability as the rest of the graph.
\textbf{Table ~\ref{table:synthetic}} summarizes these parameter choices.

\begin{table}[htb!]
\centering
\begin{adjustbox}{width=1\textwidth}
\begin{tabular}{@{}lll@{}}
\toprule
\textbf{Variable}               & \textbf{Values}                            & \textbf{Description}\\ \midrule
Motif type             & barbell, wheel, random, star, clique & Topology of planted motif                                               \\
Motif size             & 3, 5, 10, 20                  & Number of nodes in planted motif                                        \\
Concentration          & 30\%, 50\%, 100\%             & Fraction of graphs where motif appears.                                 \\
Number of motifs       & 1, 3, 10                      & Number of motifs in single dataset.                                     \\
Distortion probability & .01, .02, .05       & Probability of distorting an edge in motif.                             \\
\bottomrule
\end{tabular}
\end{adjustbox}
\caption{Synthetic dataset construction parameters}
\label{table:synthetic}
\end{table}

\subsection{Adapting EdgePool for motif mining}

The EdgePool layer introduced in ~\cite{diehl2019edge} computes a new graph $G'$ from an input graph $G = (V, E, \bf X)$ by first computing a score $s_{uv} = \textrm{MLP}(x_u, x_v) \quad \forall (u, v) \in E$. 
A greedy algorithm chooses edges with the highest scores first and collapses them into a new node we denote as $uv$, and computes a new node embedding $x_{uv} = \textrm{Pool}(x_u, x_v)$.
The Pool operation is typically a sum pool but can be any differentiable operation.
Contraction is stopped when no edges that connect previously pooled nodes are left.
Nodes that do not belong to a contracted edge are assigned to a new node in $G'$.
That is, after the pooling step, every node in $G$ is assigned to exactly one node in $G'$ and contracted pairs of nodes are mapped to the same node in $G'$.
In \algoname we wish to allow for the possibility that some graphs do not give rise to motifs if they do not fit the concentration criteria.
For this reason we replace the greedy contraction algorithm with a random sampling.
Each edge is assigned a probability using a Sigmoid layer and the new node set is computed by iterating through each edge and adding it to the pooled edges with probability proportional to $s(u, v)$.
As such, our version of coarsening is stochastic.
Additionally, EdgePool computes joint embeddings $x_{uv}$ only after deciding to contract an edge.
Instead we wish for the model to make the pooling decision as a function of the \emph{joint} subgraph and thus in our modification, $s(u, v) = \textrm{Sigmoid} (\textrm{Pool}(x_u, x_v))$,
for which joint embeddings are computed before the merging step.

\subsection{\algoname architecture}

The models built for main results follow the hyperparameters choices outlined in \textbf{Table ~\ref{table:hyper}}. Once a model is trained, the LSH-based decoding phase admits two choices. The first is the dimensionality of the hash digest which we vary from $\{8, 16, 32\}$ and which controls the collision probability when assigning embeddings to a motif label. 
Larger hash digests have lower collision probabilities and are therefore more sensitive to variability in motif structure.
The second choice in decoding is the \algoname pooling layer to use, for larger motifs we choose higher layers. This is also varied from 2 to 4 in our experiments.

\begin{table}[htb!]
\centering
\begin{tabular}{@{}ll@{}}
\toprule
Parameter       & Values                                        \\ \midrule
Embedding model & $\textrm{MLP}$, $\textrm{ReLU}$ activation             \\
Embedding size  & 8                                             \\
Pooling layers  & 4                                             \\
Scoring model   & $\textrm{MLP}$, $\textrm{Sigmoid}$ activation \\
$\lambda$       & 1                                             \\
$\beta$         & 1                                             \\ \bottomrule
\end{tabular}
\caption{Hyperparameter choices for \algoname}
\label{table:hyper}
\end{table}

\subsection{Observed runtime}

\begin{figure}
    \centering
    \includegraphics[width=.7\textwidth]{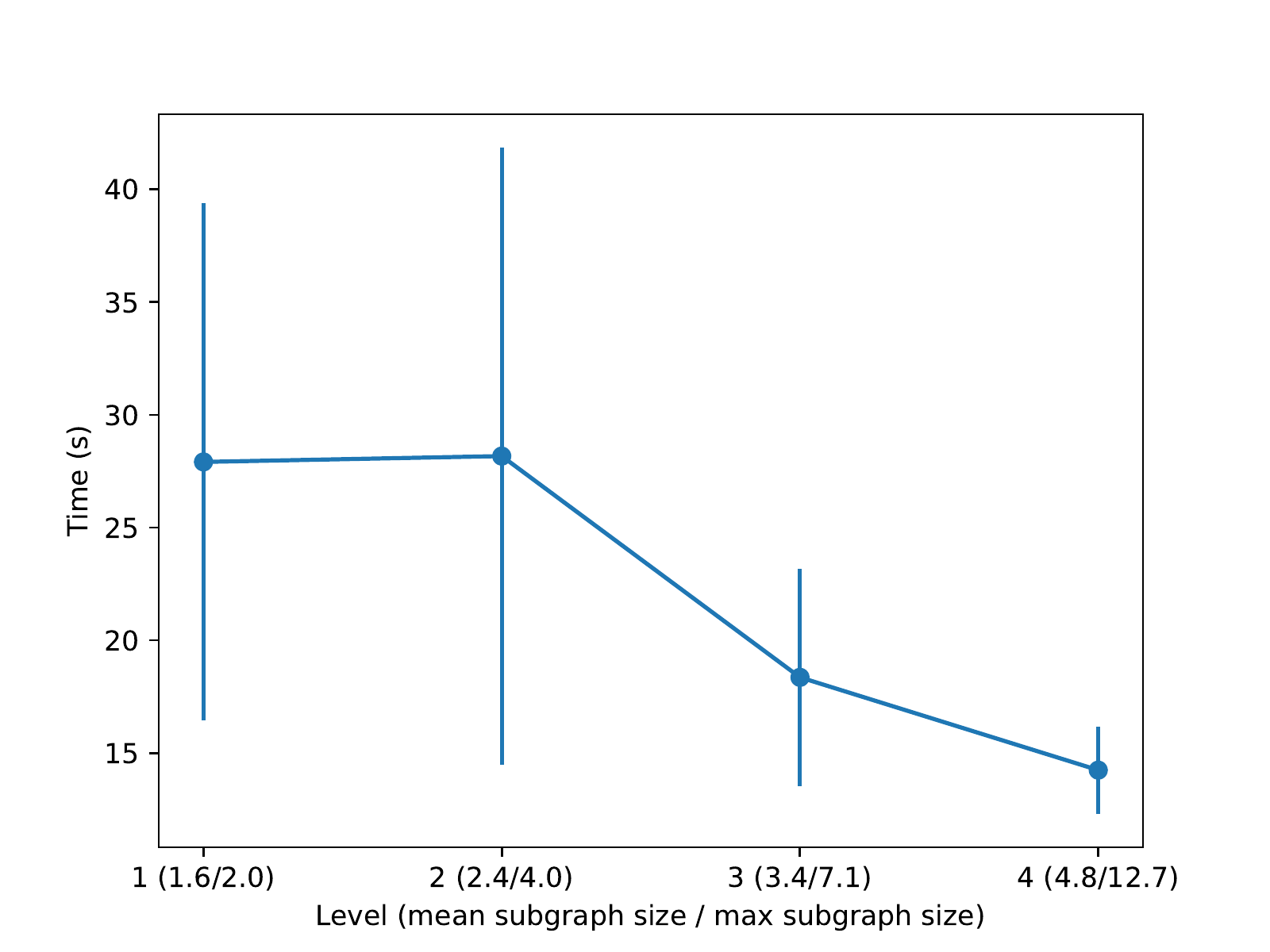}
    \caption{Decoding Runtime. Given a trained model, we decode motifs at each specified pooling layer (1-4, on x-axis) and note the resulting mean subgraph size and largest subgraph found at each layer.}
    \label{fig:times}
\end{figure}

Training time on typical synthetic runs using 1 NVIDIA GeForce GTX 1080 varied between 4-8 hours on the synthetic datasets.
We benchmark the decoding time on one of our synthetic datasets of 1000 graphs of 20 nodes each.
Results for 30 runs for decoding at 1 to 4 pooling layers are shown in \textbf{Figure ~\ref{fig:times}}.
Decoding was performed on a personal laptop using 1.6GHz dual-core Intel Core i5 processor. 

\subsection{Comparison with exact method}

As an external comparison of motif retrieval, we compare with the popular exact motif mining algorithm, MFinder ~\cite{kashtan2004efficient}.
Since this method enumerates substructures the software crashed after 5 hours on datasets containing motifs of size 5 which is the smallest motif we tested on \algoname.
Single-motif datasets with size 4 motifs did execute successfully for each of our 5 motif topologies, and 4 distortion levels.
Resulting M-Jaccard values are summarized in \textbf{Table~\ref{table:mfinder}}.
\mfinder is unable to retrieve the planted motif for all topologies except for cliques which are almost perfectly recovered.

\begin{table}
\centering
\begin{tabular}{lrrrr}
\toprule
{} & \multicolumn{4}{l}{} \\
 &      $d=0.00$ & $d=0.01$ & $d=0.02$ & $d=0.05$ \\
 &           &      &      &      \\
\midrule
\textbf{barbell   } &      0.00 & 0.04 & 0.08 & 0.22 \\
\textbf{clique    } &      1.00 & 0.94 & 0.88 & 0.73 \\
\textbf{random    } &      0.00 & 0.05 & 0.09 & 0.18 \\
\textbf{star      } &      0.00 & 0.10 & 0.18 & 0.33 \\
\bottomrule
\end{tabular}
\caption{M-Jaccard coefficient using MFinder \cite{kashtan2004efficient}} on synthetic dataset with motifs of size 4.
\label{table:mfinder}
\end{table}

Because in \mfinder, only isomorphic subgraphs are counted as part of a motif, we observe that when the motif is missed, it is completely missed (M-Jaccard $\rightarrow$ 0).
This is particularly noticeable when distortion probability is close to zero, and the opposite trend can be seen for cliques where the algorithm successfully detects the motif it catches all instances and performance decreases with distortion probability.

\end{document}